\documentclass[runningheads]{llncs}
\usepackage{graphicx}
\usepackage[caption=false]{subfig}
\usepackage{tabularx,calc,array,multirow,makecell, enumitem, amsfonts}

\usepackage{textcomp, gensymb} 
\graphicspath{ {./images/} }

\usepackage{tikz}
\usepackage{comment}
\usepackage{amsmath,amssymb} 
\usepackage{color}

\usepackage[accsupp]{axessibility}  

\usepackage[width=122mm,left=12mm,paperwidth=146mm,height=193mm,top=12mm,paperheight=217mm]{geometry}

\begin{document}
\pagestyle{headings}
\mainmatter
\def\ECCVSubNumber{5}  

\title{RelMobNet: End-to-end relative camera pose estimation using a robust two-stage training} 

\titlerunning{RelMobNet}
%
\author{Praveen Kumar Rajendran\inst{1} \and
Sumit Mishra\inst{2} \and
Luiz Felipe Vecchietti\inst{3}\and
Dongsoo Har\inst{4}}

%
\authorrunning{P. Rajendran et al.}
%
\institute{Division of Future Vehicle, KAIST, South Korea \and
The Robotics Program, KAIST, South Korea
\and
Data Science Group, Institute for Basic Science, South Korea\\ \and
The CCS Graduate School of Mobility, KAIST, South Korea
}
\maketitle

\begin{abstract}
Relative camera pose estimation, i.e. estimating the translation and rotation vectors using a pair of images taken in different locations, is an important part of systems in augmented reality and robotics. In this paper, we present an end-to-end relative camera pose estimation network using a siamese architecture that is independent of camera parameters. The network is trained using the Cambridge Landmarks data with four individual scene datasets and a dataset combining the four scenes. To improve generalization, we propose a novel two-stage training that alleviates the need of a hyperparameter to balance the translation and rotation loss scale. The proposed method is compared with one-stage training CNN-based methods such as RPNet and RCPNet and demonstrate that the proposed model improves translation vector estimation by 16.11\%, 28.88\%, and 52.27\% on the Kings College, Old Hospital, and St Marys Church scenes, respectively. For proving texture invariance, we investigate the generalization of the proposed method augmenting the datasets to different scene styles, as ablation studies, using generative adversarial networks. Also, we present a qualitative assessment of epipolar lines of our network predictions and ground truth poses.
\end{abstract}

\section{Introduction}

Recently, image-based localization, i.e. the process of determining a location from images taken in different locations, has gained substantial attention. In particular, estimating the relative camera pose, i.e. the camera's location and orientation in relation to another camera's reference system, from a pair of images is an inherent computer vision task that has numerous practical applications, including 3D reconstruction, visual localization, visual odometry for autonomous systems, augmented reality, and virtual reality \cite{kim2020pose,melekhov2017relative}. Pose accuracy is critical to the robustness of many applications \cite{kendall2015posenet}. Traditionally, determining the relative pose entails obtaining an essential matrix for calibrated cameras \cite{hartley1997defense}. Uncalibrated cameras, on the other hand, necessitate fundamental matrix estimation which captures the projective geometry between two images from a different viewpoint \cite{hartley2013multiple}.

\begin{figure}[tb]
     \centering
     \subfloat[]{\includegraphics[width=.24\textwidth,height=\textheight,keepaspectratio]{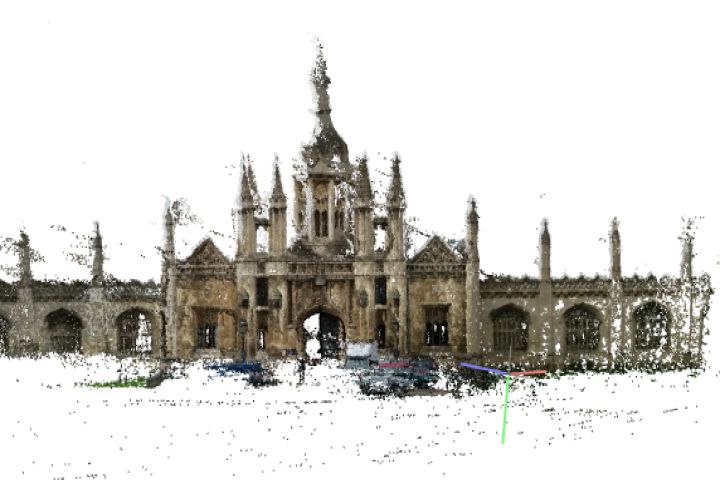}\label{value1}}
     \subfloat[]{\includegraphics[width=.24\textwidth,height=\textheight,keepaspectratio]{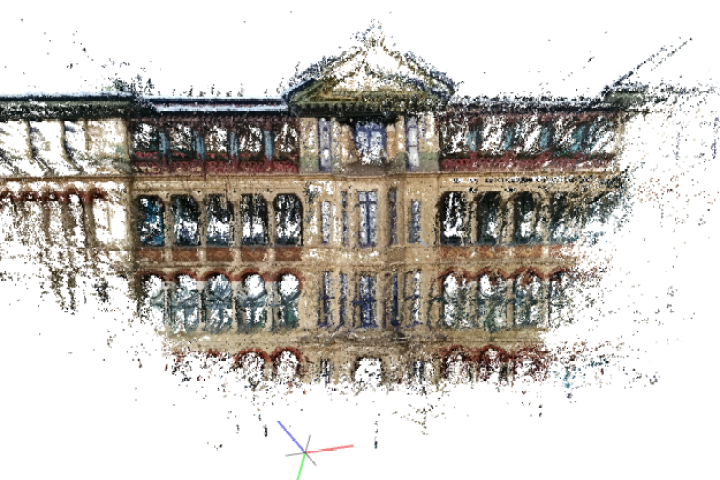}\label{value2}}
     \subfloat[]{\includegraphics[width=.24\textwidth,height=\textheight,keepaspectratio]{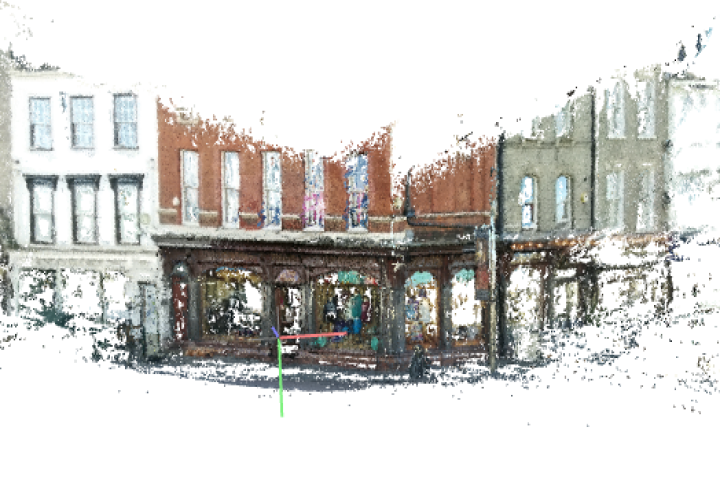}\label{value3}}
     \subfloat[]{\includegraphics[width=.24\textwidth,height=\textheight,keepaspectratio]{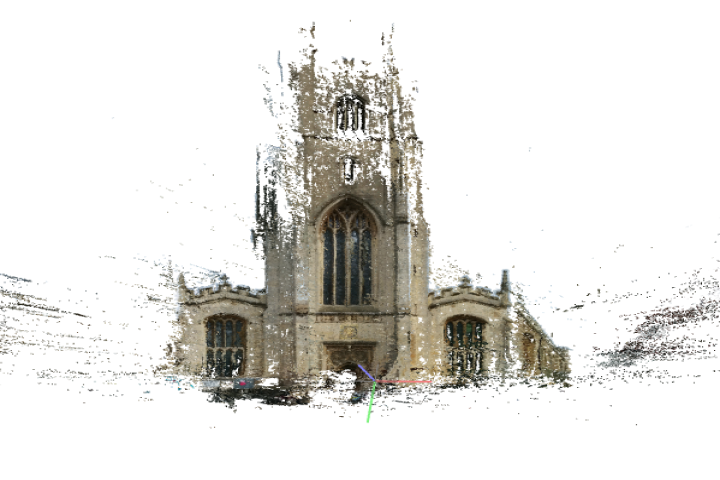}\label{value4}}
     \setlength{\belowcaptionskip}{-8pt}
     \caption{Dense 3-dimensional reconstruction for different scenes in the Cambridge Landmarks dataset, (a) \textit{Kings College}(seq2), (b) \textit{Old Hospital}(seq3), (c) \textit{Shop Facade}(seq3), and (d) \textit{St Marys Church}(seq3), using the COLMAP \cite{schoenberger2016sfm} algorithm}
     \label{fig:cambridge}
\end{figure}

Classical methods, such as SIFT \cite{lowe2004distinctive}, SURF \cite{bay2008speeded}, and FAST \cite{philbin2007object}, involve extracting key points for feature matching between two images. The relative pose is then estimated by exploiting 3D geometry properties of the extracted features \cite{bay2008speeded,lowe2004distinctive,philbin2007object,rublee2011orb}. However, they can suffer from inconsistent feature correspondences due to reasons such as insufficient or repetitive patterns, narrow overlap, large scale change or perspective change, illumination variation, noise, and blur \cite{chen2021wide}. Among multiple feature-extractors, SIFT and SURF detectors are more resistant to scale differences but are relatively slow. Multiple approaches to improve the performance of key point based methods using non-maximal suppression were proposed \cite{bailo2018efficient}. Nonetheless, these approaches use these key points to estimate a relative pose, where the estimated translation vector is up to a scale factor, i.e. a proportional vector to a translation vector. Pose estimation quality greatly relies on corresponding features, which are traditionally extracted by different key point extractors, and it should be noted that each feature extractor has a specific resilience for speed, light condition, and noise. However, an ideal key point extractor must be invariant and robust to various lighting conditions and transformations at the same time. In this paper, we investigate an end-to-end learning-based method with the aim of addressing these challenges.

Machine learning approaches have achieved satisfactory results in various tasks, including recharging sensor networks \cite{moraes2017distributed}, wireless power transfer \cite{hwang2018ferrite}, power grid operation \cite{lee2019optimal}, and robotic control \cite{seo2019rewards}. Recently, several deep learning models have been investigated based on the success of deep convolutional neural network (CNN) architectures for images. CNNs achieve high performance on computer vision tasks such as image classification, object detection, semantic segmentation, place recognition, and so on. Here, we investigate a computer vision task knows as camera pose estimation. To tackle this problem, firstly, PoseNet was proposed as a pose regressor CNN for the camera localization task \cite{kendall2015posenet}. Advancing over PoseNet, various CNN-based approaches have been proposed for the relative pose estimation. Generally, all of these methods handle this as a direct regression task such as in RPNet \cite{en2018rpnet}, cnnBspp \cite{melekhov2017relative}, and RCPNet \cite{yang2020rcpnet}. The RPNet algorithm uses a manual hyperparameter for weighing between the translation and rotation losses. This hyperparameter is tuned for models that are trained for different scenes, i.e. each model uses a different value depending on which dataset it is trained on. RCPNet. on the other hand, uses a learnable parameter to balance between these losses. Also, the aforementioned methods crop the images to a square of the same size as that of the pre-trained CNN backbone. DirectionNet \cite{chen2021wide} uses a different parameterization technique to estimate a discrete distribution over camera pose. As identified by previous works, CNNs can produce good and stable results where traditional methods might fail \cite{kendall2015posenet}. Deriving on the benefits from previous CNN-based methods, we propose yet another end-to-end approach to tackle the relative pose regression problem using a siamese architecture using a MobileNetV3-Large backbone to produce entire translation and rotation vectors. Furthermore, we discuss a simple and compact pipeline using COLMAP \cite{schoenberger2016sfm} to gather relative pose data with mobile captured videos, that also can be applied to visual odometry.

In this paper, our key contributions can be outlined as follows
\begin{enumerate}
    \item We present RelMobNet, a siamese convolutional neural network architecture using a pre-trained MobileNetV3-Large backbone with shared weights. The proposed architecture overcome the necessity of using input images with the same dimensions as the one used by the pre-trained backbone by exploiting the use of adaptive pooling layers. In this way, the images can maintain their original dimensions and aspect ratio during training.
    \item With the help of a novel two-stage training procedure, we alleviate the need of an hyperparameter to weight between translation and rotation losses for different scenes. The translation loss and the rotation loss are given the same importance during training which eliminates the need of a grid search to define this hyperparameter for each scene.
    \item For proving texture invariance, we investigate the generalization of the proposed method augmenting the datasets to different scene styles using generative adversarial networks. We also present a qualitative assessment of epipolar lines of the proposed method predictions compared to ground truth poses.
\end{enumerate}
The rest of this paper is structured as follows. Section 2 discusses related work. Section 3 presents the dataset used in the experiments and detail the proposed network architecture and training process. The experiments and the analysis, both qualitative and quantitative, are presented in Section 4. Section 5 concludes this paper.

\section{Related Work}

\subsection{Feature correspondence methods}

Methods based on SIFT, SURF, FAST, and ORB feature detectors are viable options for solving the relative pose estimation. These methods take advantage of the epipolar geometry between two views, which is unaffected by the structure of the scene \cite{hartley2013multiple}. Conventionally, the global scene is retrieved using keypoint correspondence by calculating the essential matrix which uses RANSAC, an iterative model conditioning approach, to reject outliers \cite{hartley1997defense,nister2004efficient,fischler1981random}. Matching sparse keypoints between a pair of images using those descriptors can help unearth the relative orientation by the following pipeline: i) keypoint detection; ii) computation of local descriptors; and iii) matching local descriptors between an image pair. Even so, the performance still depends on correct matches and speculation of the textures in the given images \cite{raguram2008comparative}. Inaccuracy in feature matching due to insufficient overlap of a given image pair can have a significant impact on the performance.

Several deep learning-based methods are proposed to tackle sub-issues in the traditional pipeline using feature correspondence methods. In \cite{poursaeed2018deep}, a method is proposed for improving fundamental matrix estimation using a neural network with specific modules and layers to preserve mathematical properties in an end-to-end manner. D2-Net and SuperGlue are proposed to tackle feature matching \cite{dusmanu2019d2,sarlin2020superglue}. D2-Net reported promising results leveraging a CNN for performing two functions of detection and description of the keypoints by deferring the detector stage to utilize a dense feature descriptor \cite{dusmanu2019d2}. SuperGlue used a graph neural network to match sets of local features by finding correspondence and rejecting non-matchable points simultaneously \cite{sarlin2020superglue}. Differentiable RANSAC (DSAC) was proposed to enable the of use robust optimization used in deep learning-based algorithms by converting non-differentiable RANSAC into differentiable RANSAC \cite{fischler1981random,brachmann2017dsac}. Detector-free local feature matching with transformers (LoFTR) employs a cross attention layer to obtain feature descriptors conditioned on both images to obtain dense matches in low texture areas where traditional methods struggle to produce repeatable points \cite{sun2021loftr}.

\subsection{End-to-End methods}
PoseNet \cite{kendall2015posenet} is the first end-to-end pose regressor CNN that can estimate 6 DOF camera pose with an RGB image. PoseNet fits a model on a single landmark scene to estimate an absolute pose from an image. The method is a robust alternative to low and high lighting and motion blur by learning the scene implicitly. However, PoseNet is trained to learn a specific scene, which makes it difficult to scale for other scenes. In \cite{kendall2017geometric} a novel geometric loss function for improving the performance across datasets by leveraging properties of geometry and minimizing reprojection errors is proposed.

To solve the relative pose problem, firstly, \cite{melekhov2017relative} demonstrated that the estimation of the relative pose can be made by using image pairs as an input to the neural network. RPNet \cite{en2018rpnet} proposed to recover the relative pose using the original translation vector in an end-to-end fashion. In RPNet the image pairs are generated by selecting eight different images in the same sequence. RCPNet \cite{yang2020rcpnet} utilizes a similar approach to interpret relative camera pose applied to the autonomous navigation of Unmanned Aerial Vehicles (UAVs). RCPNet presented their results for a model trained with multiple scenes reporting comparative results with PoseNet and RPNet. All of the aforementioned approaches used Euclidean distance loss with a hyperparameter to balance between translation and rotation losses. This hyperparameter was tuned differently for models trained for different scenes. In contrast to these approaches, in this paper we show that, without a scene-dependent hyperparameter to balance between these two losses, it is possible to estimate translation vectors with higher accuracy compared to the previous methods. We also show that we can feed and exploit full image details (without cropping to the same input size of the pretrained network) as we can make use of CNNs invariance to size property \cite{graziani2021scale}. Using adaptive average pooling layers \cite{adaptiveavgpool2d-pytorch1.12documentation} before dense layers ensures that the output activation shape matches the desired feature dimension of the final linear layers, and thus ensuring the ability to handle flexible input sizes.

\begin{figure}[tb]
\captionsetup[subfigure]{labelformat=empty}
\includegraphics[width=\textwidth,height=\textheight,keepaspectratio]{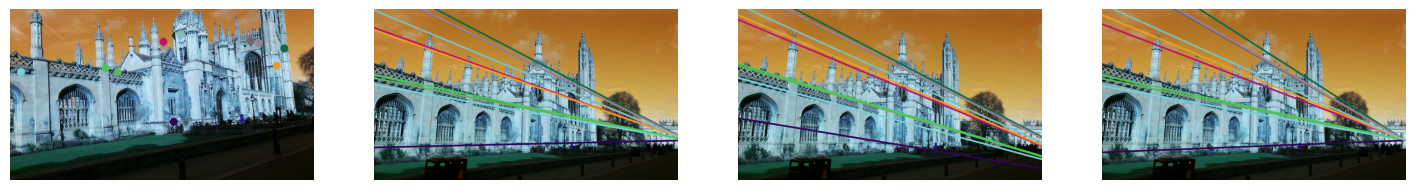}
\includegraphics[width=\textwidth,height=\textheight,keepaspectratio]{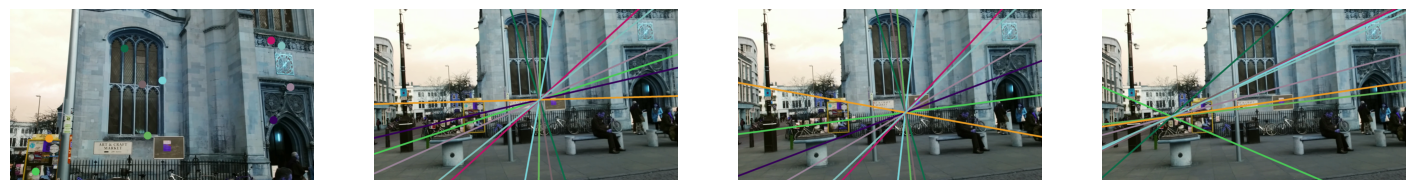}
\includegraphics[width=\textwidth,height=\textheight,keepaspectratio]{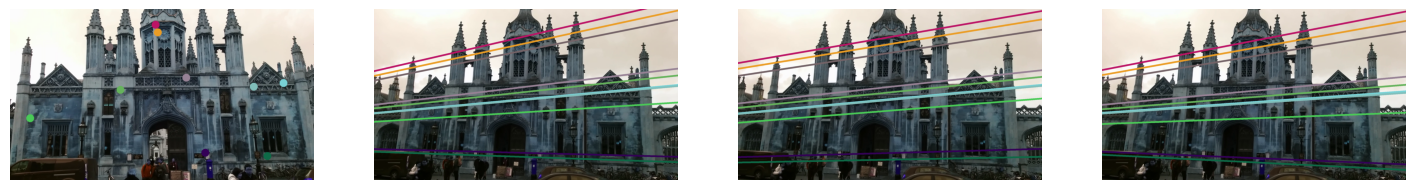}
\subfloat[Reference image \qquad \quad GT Pose \quad\quad\qquad\quad SIFT+LMedS \qquad \quad Predicted Pose]{\includegraphics[width=\textwidth,height=\textheight,keepaspectratio]{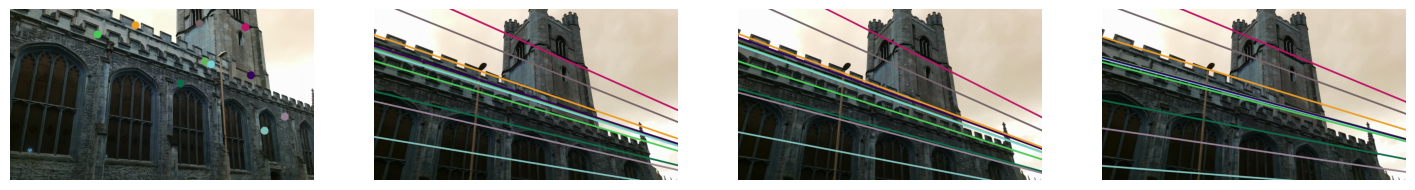}}
\setlength{\belowcaptionskip}{-8pt}
\caption{Qualitative evaluation of epipolar lines for corresponding key-points of reference images, as represented by same colour of lines. First column represents the reference images with keypoints. Second, third, and fourth column represents epipolar lines based on, ground truth pose, SIFT+LMeds, and proposed RelMobnet}\label{fig:epilines}
\end{figure}

\section{Methodology}

\begin{figure}[tb]
    \includegraphics[width=\textwidth,height=\textheight,keepaspectratio]{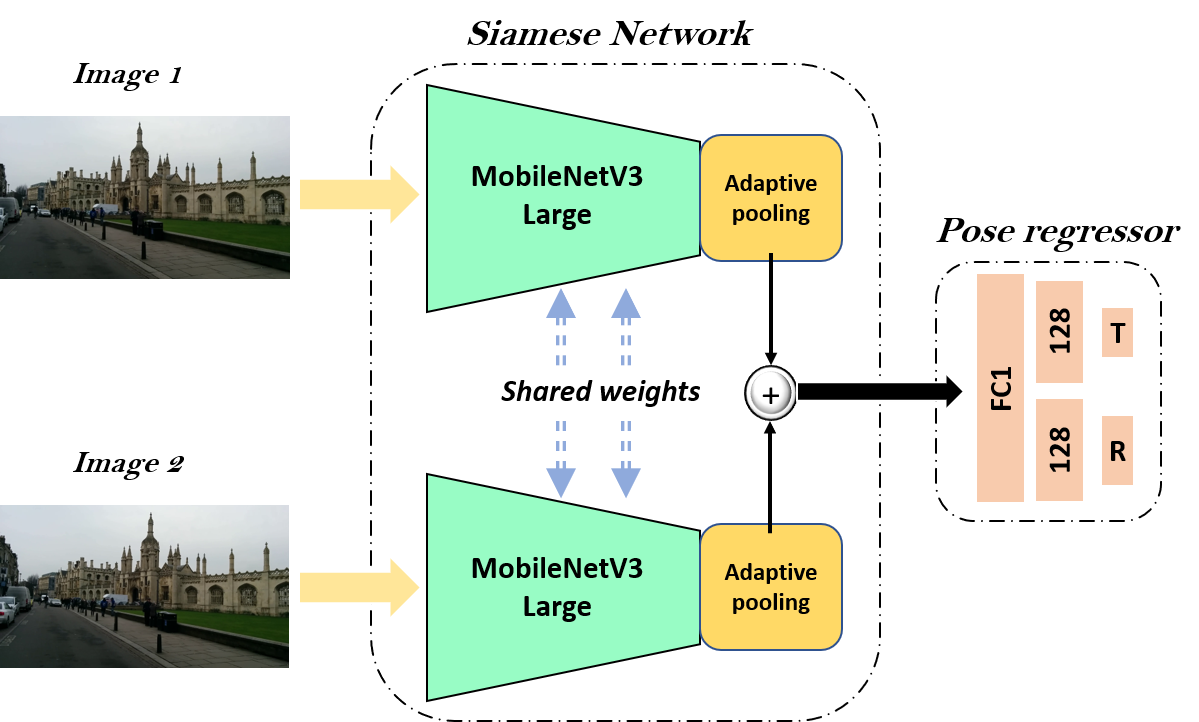}
    \setlength{\belowcaptionskip}{-8pt}
    \caption{RelMobNet: a siamese convolutional neural network architecture using pre-trained MobileNetV3-Large backbones with adaptive average pooling layers. The output of the parallel branches in the siamese network are concatenated and with a pose regressor to estimate a translation and a rotation vector. The adaptive pooling layers are added to handle variable input image sizes
    }\label{fig:arcitecture}
\end{figure}

\subsection{Dataset image pairs}
The Cambridge Landmarks dataset contain video frames with their ground truth absolute camera pose for outdoor re-localisation as presented in \cite{kendall2015posenet}. The dataset is constructed with the help of structure from motion (SFM) methods. We visualized a few sequence of scenes using COLMAP as shown in Fig.~\ref{fig:cambridge}. Relative pose estimation pairs are used as that of RPNet \cite{en2018rpnet}. For example, the images from the same sequence of video frames from a single scene are used to make a pair. The camera's relative pose is expressed as a 3-dimensional vector for translation \(t = (x, \thinspace y, \thinspace z)\) and a 4-dimensional vector for rotation \(q = (qw, \thinspace qx, \thinspace qy, \thinspace qz)\), i.e. \((qw + \textbf{i} \thinspace qx + \textbf{j} \thinspace qy + \textbf{k} \thinspace qz)\) where \(\textbf{i, j, k}\) represent the imaginary part of a quaternion. For unit quaternions \( q_1, \thinspace q_2 \) equivalent rotation matrices are defined as \( R_1, \thinspace R_2 \). Respective translations are represented as \(t_1, \thinspace t_2\). A projection matrix is used to project a point from a reference global coordinate system to the camera coordinate system and is composed by a rotation matrix and a translation vector. Relative translation, and relative rotation, both in matrix and in quaternion form,  are represented as \(t_{rel},\thinspace R_{rel}, \thinspace q_{rel}\), respectively. The conjugate of a quaternion is represented by $q^*$. The calculation of \(q_{rel}\) and \(t_{rel}\) is given as follows.
\begin{equation}
q_{rel} = q_2 \times q_1^*
\end{equation}
\begin{equation}
t_{rel} = R_2(-R_1^T t_1) + t_2
\end{equation} 

The ground truth label for rotation is based on a normalized quaternion which is used to train the deep learning model. When the deep learning model output the vector for rotation, it is normalized to unit length at test evaluation. As a novel training procedure, we perform the model training in two-stages, using two sets of ground truth. In the \textit{first set}, both rotation and translation values are normalized. In the \textit{second set}, only rotation values are normalized and translation values are unnormalized.

\subsection{Architecture Details}

As shown in Fig.~\ref{fig:arcitecture}., to compute the features from two images, a siamese model composed of two parallel branches, each one with a MobileNetV3-Large architecture that share weights is utilized. The MobileNetV3-Large architecture is chosen as our CNN architecture because of its computational efficiency obtained by redesigning expensive intermediate layers with the Neural Architecture Search (NAS) method \cite{howard2019searching}. In addition, we add adaptive average pooling layer, implemented as in the PyTorch library \cite{NEURIPS2019_9015}  to handle input images with variable sizes. In our architecture, the layer before the adaptive pooling layer gives 960 channels with different image features. After these channels pass to the 2-dimensional adaptive pooling layer, the output provides a single (1*960-dimensional) vector.

The outputs from the two parallel branches are then concatenated to make up a flattened feature vector. This feature vector is the input of the pose regressor block. The first layer in the pose regressor block consists of 1920 units of neurons to accommodate for the concatenated feature vector. Secondly, it has two parallel 128-unit dense layers to predict the relative translation \((x, \thinspace y, \thinspace z)\) and the rotation \((qw, \thinspace qx, \thinspace qy, \thinspace qz)\), respectively. Lastly, Euclidean distance is set as the objective criterion for translation and rotation to train our network.
Inspired by previous works \cite{sarlin2020superglue,yew2022regtr}, we investigated the effectiveness of cross-attention layers after feature extraction. Results, however, were not superior to the plain siamese architecture.

In our architecture, we do not employ a hyperparameter to balance between the rotation and translation losses, i.e. translation and rotation losses are given the same importance. Defining \(\overline{rpose}\) as the ground truth and \(\widehat{rpose}\) as the prediction of the network, the loss is given as
\begin{equation}
\begin{split}
\mathcal{L} (\widehat{rpose}, \overline{rpose}) = \sum_{i}^{n} (\parallel \hat{t}_{rel}^i - {t}_{rel}^i\parallel_2 + \parallel \hat{q}_{rel}^i - {q}_{rel}^i\parallel_2 )
\end{split}
\end{equation} 

where \(n\) denotes the number of samples in the current batch, \(\hat{t}_{rel}\) is the predicted relative translation and  \(\hat{q}_{rel}\) is the predicted relative rotation respectively.

\section{Experiments and Analysis}

\subsection{Baseline Comparison}

For comparison, the findings of Yang et al \cite{yang2020rcpnet} is adopted. With the proposed architecture, we trained a shared model combining multiple scenes into a single dataset and individual models on individual scene datasets similarly to RCPNet \cite{yang2020rcpnet}. We combine four scenes \emph{Kings College, Old Hospital, Shop Façade, St Mary’s Church} of the Cambridge Landmarks dataset for training whereas RCPNet kept \textit{Shop Facade} scene unseen at the training phase. We used the training and test splits as that of the RPNet \cite{en2018rpnet}. For the shared model, we integrated all of the training splits from four landmarks datasets into a single training set to train the CNN.

\textbf{Two-stage training. }The training is performed in two-stages, In the first stage, we used the unit quaternion and a normalized version of the translation vector as the ground truth (i.e, \textit{first set} ground truth labels) and trained it for 30 epochs. In the second stage, we finetuned the first stage model with the same training images for 20 epochs with the unit quaternion and an unnormalized translation vector as the ground truth (i.e, \textit{second set} ground truth labels). By performing model training in this way, the model was able to learn the balance between the rotation and translation vector implicitly and we observed a faster convergence of the model. 


\begin{table}[tb]
\begin{center}
\setlength{\belowcaptionskip}{-8pt}
\caption{Performance comparison on four scenes of the Cambridge Landmarks dataset adopted from \cite{yang2020rcpnet}. \textbf{Bold} numbers represent results with better performance (median values are presented)\\
\textbf{Note}: Our Two-stage model results are compared with baselines.(trained with \textit{first set}, and \textit{second set} ground truths)
$\rightarrow$ 4 Indvidual scene models, 1 Shared scence model}
\label{table:perf}
\resizebox{\columnwidth}{!}{\begin{tabular}{|c|c|c|c|c|c|c|c|c|c|c|c|}
\hline
\multirow{2}{*}{ Scene } & \multicolumn{2}{c}{Frames} & \multicolumn{2}{c}{Pairs} & \multirow{2}{*}{ \makecell{Spatial \\ Extent(m)} } & \multirow{2}{*}{ RPNet } & \multicolumn{2}{c}{RCPNet} & \multicolumn{2}{c}{Ours} & (Shared) \\
& Test & Train & Test & Train & & & (Individual) & (Shared) & (Individual) & (Shared) & \text{\%} Change in Translation \\
\hline
King’s C. & 343 & 1220 & 2424 & 9227 & 140×40 & 1.93m,3.12$^{\circ}$ & 1.85m,\textbf{1.72}$^{\circ}$ & 1.80m,1.72$^{\circ}$ & \textbf{1.45m},2.70$^{\circ}$ & 1.51m,2.93$^{\circ}$ & 16.11 \\
Old Hospital & 182 & 895 & 1228 & 6417 & 50×40 & 2.41m,4.81$^{\circ}$ & 2.87m,\textbf{2.41}$^{\circ}$ & 3.15m,3.09$^{\circ}$ & 2.51m,3.60$^{\circ}$ & \textbf{2.24m},3.63$^{\circ}$ & 28.88 \\
St Mary’s C. & 530 & 1487 & 3944 & 10736 & 80x60 & 2.29m,\textbf{5.90}$^{\circ}$ & 3.43m,6.14$^{\circ}$ & 4.84m,6.93$^{\circ}$ & \textbf{2.14m},6.47$^{\circ}$ & 2.31m,6.30$^{\circ}$ & 52.27 \\
Shop Facade & 103 & 231 & 607 & 1643 & 35x25 & 1.68m,7.07$^{\circ}$ & 1.63m,7.36$^{\circ}$ & {\makecell{13.8m,28.6$^{\circ}$ \\ (unseen)}} & 2.63m,11.80$^{\circ}$ & \textbf{1.34m,5.63}$^{\circ}$ & Not applied \\
\hline
\end{tabular}}
\end{center}
\end{table}

\begin{figure}[ht]
     \centering
     \subfloat[]{\includegraphics[width=.48\textwidth,height=\textheight,keepaspectratio]{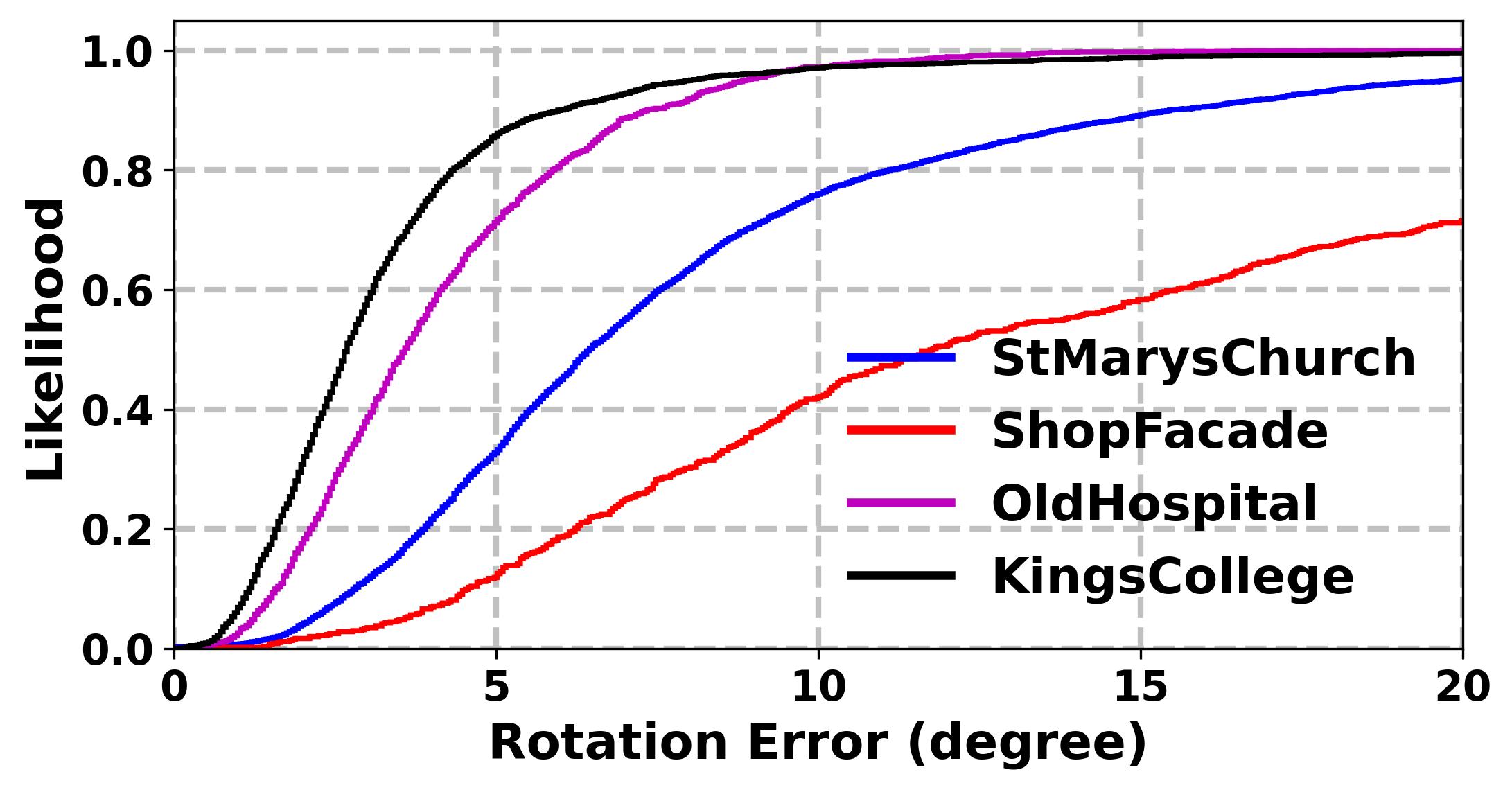}\label{a}}
     \subfloat[]{\includegraphics[width=.48\textwidth,height=\textheight,keepaspectratio]{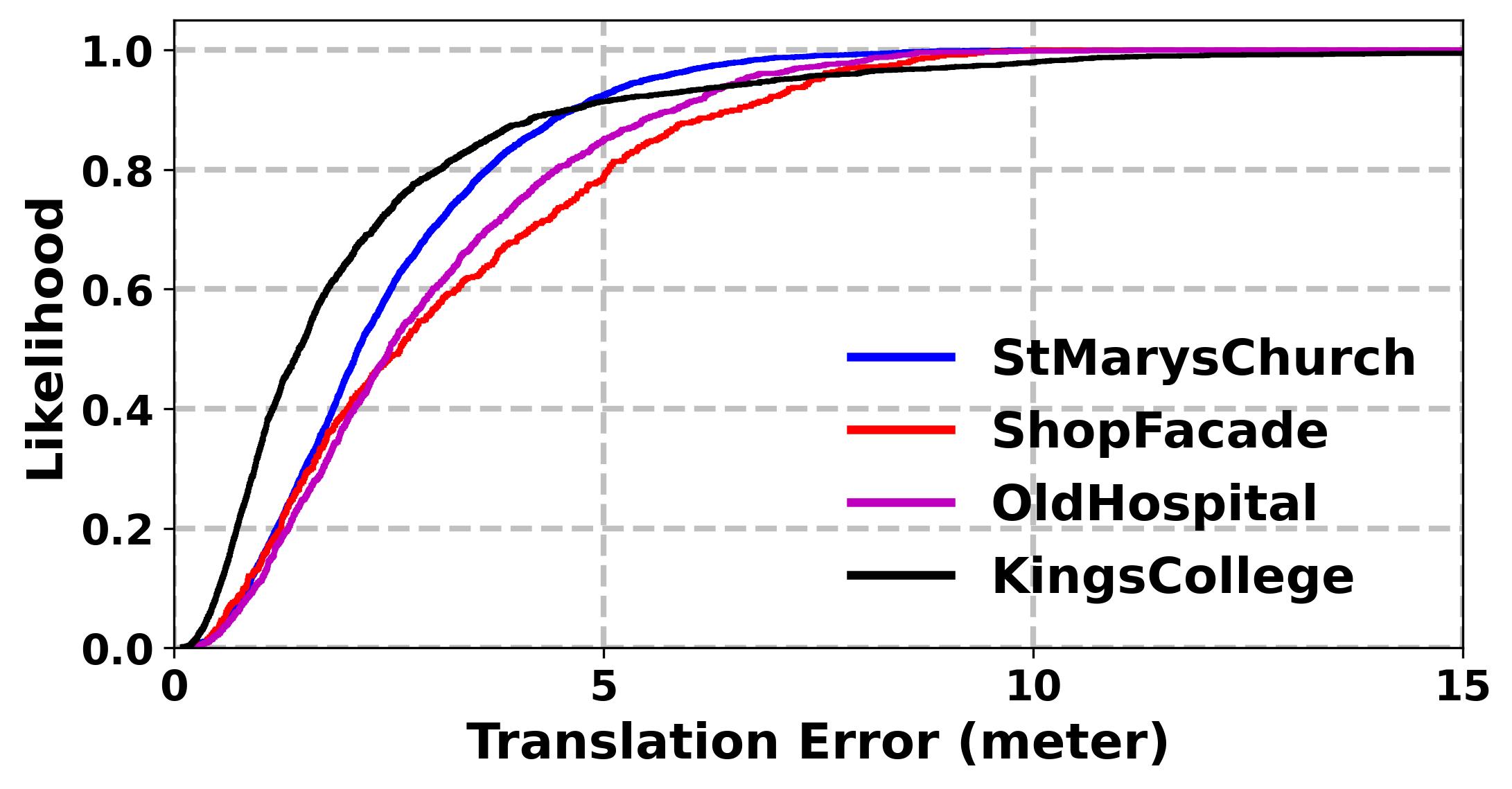}\label{b}} \\
     \subfloat[]{\includegraphics[width=.48\textwidth,height=\textheight,keepaspectratio]{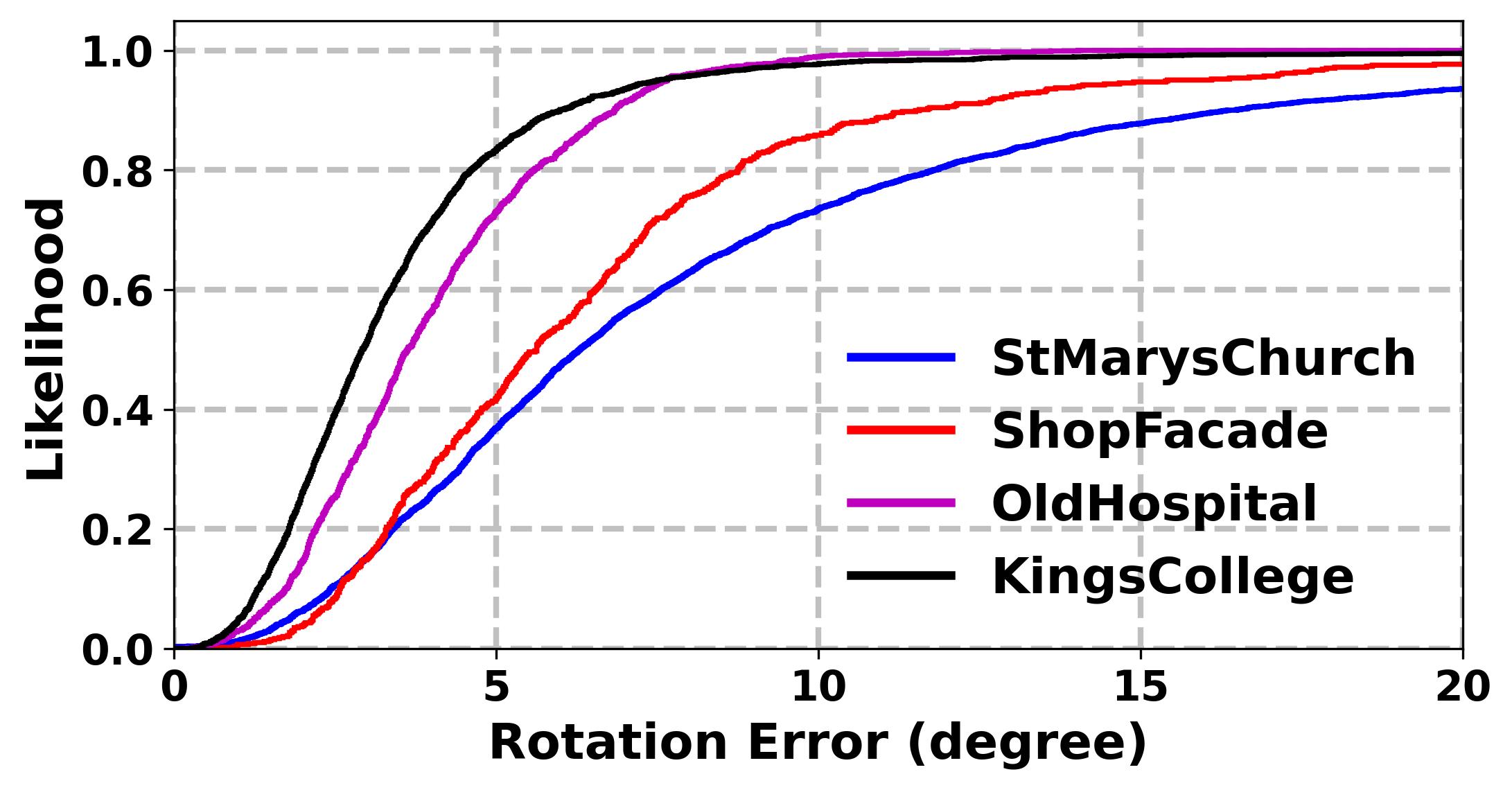}\label{c}}
     \subfloat[]{\includegraphics[width=.48\textwidth,height=\textheight,keepaspectratio]{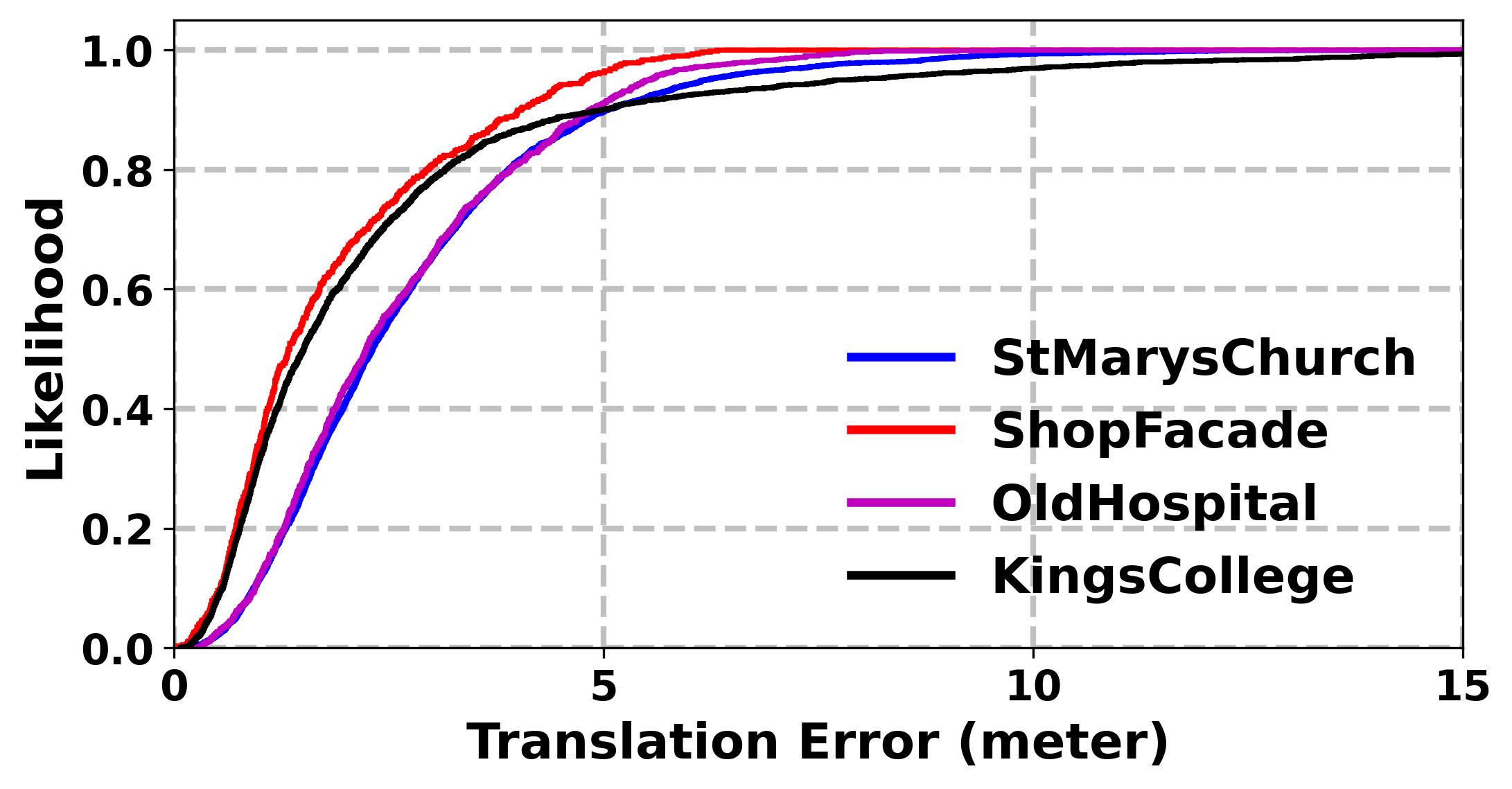}\label{d}}
     \caption{Cumulative histogram of test errors in rotation and translation for individual models and the shared model. The higher the area under the curve, the better the estimation. (a) rotation error using individual models (b) translation error using individual models (c) rotation error using the shared model (d) translation error using the shared model }
     \label{fig:histogram}
\end{figure}

As the baseline methods train separate models for each scene, we also train individual models for comparison. The same two-stage training as explained above is used for individual training. The input image was downsized with the height set to 270 pixels while maintaining the original aspect ratio. The models are implemented in PyTorch. All models are trained on a NVIDIA 1080Ti GPU with 12GB of RAM. During training the learning rate is set to 0.001 using the ADAM optimizer \cite{kingma2014adam}. The batch size is set to 64. Results are reported in Table~\ref{table:perf}. It can be observed in Table~\ref{table:perf} that our method improve especially the translation vector predictions, achieving the best results over the base lines for the four scenes. We also show the cumulative  histogram of errors in the test sets for individual models and the shared model in Fig.~\ref{fig:histogram}. The curves if compared to first row of Figure 5 from \cite{yang2020rcpnet} are depicting the ability of the model to achieve better estimation of relative pose.

\textbf{Epipolar Lines. }For qualitative analysis, epipolar liners are drawn for selected keypoints. To plot the epipolar lines, we use a function using the OpenCV library \cite{opencv_library} which requires a fundamental matrix. The fundamental matrix \(F\) captures the projective geometry between two images from a different viewpoint. For comparison, we use feature based baseline SIFT+LMedS and the corresponding fundamental matrix \(F_{SIFT+LMedS}\) is obtained with OpenCV. For ground truth pose and the predicted pose, the fundamental matrix: \(F,\) and \(\hat F\) is calculated as follows.

\begin{equation}
F = K_{2}^{-T} \thinspace t_{rel[X]} R_{rel} \thinspace K_{1}^{-1}
\end{equation}  

\begin{equation}
\hat F = K_{2}^{-T} \thinspace \hat t_{rel[X]} \hat R_{rel} \thinspace K_{1}^{-1}
\end{equation}

where \(_{[X]}\) represents the \(skew\thinspace \thinspace symmetric\thinspace \thinspace matrix\) of a vector. \(K_{1}, \thinspace K_{2}\) are intrinsic matrices for images 1 and 2. Since we use images sampled from a video taken using the same camera we can define \(K =K_{1}=K_{2}\). As camera intrinsics \(K\) are not given in the Cambridge landmark dataset we approximate\footnote{https://github.com/3dperceptionlab/therobotrix/issues/1} K as follows  as follows, \(K \approx \begin{bmatrix}
f & 0 & c_{x}\\
0 & f & c_{y}\\
0 & 0 & 1
\end{bmatrix}\) where \(c_{x}=\) image-width / 2, \(c_{y}=\) image-height / 2, and \(f =\) image-width \(/ (tan(FOV/2) * 2)\), and \(FOV \approx 61.9\) for 30mm focal length\footnote{https://www.nikonians.org/reviews/fov-tables} is obtained based on the dataset camera focal length \cite{kendall2015posenet}. This approximation of intrinsics is acceptable as verified with COLMAP reconstruction \cite{schoenberger2016sfm} for each scene(with one sequence) in the Cambridge Landmark dataset. Fig.~\ref{fig:epilines}. demonstrates the qualitative comparison between ground truth pose, SIFT+LMedS method pose, and our predicted relative pose with the key points detected with SIFT.

\begin{figure}[tb]
     \centering
     \subfloat[]{\includegraphics[width=.48\textwidth,height=\textheight,keepaspectratio]{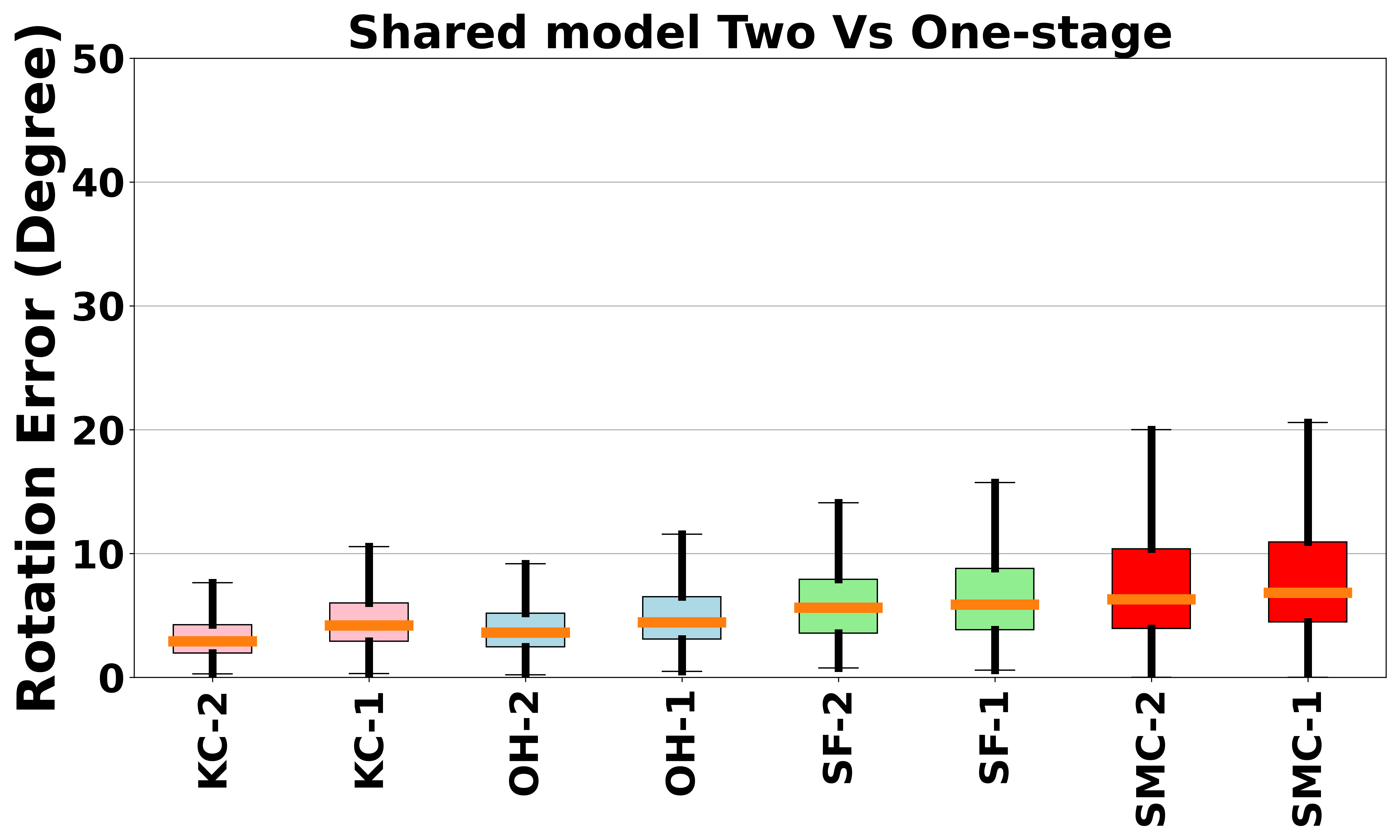}\label{sr}}
     \subfloat[]{\includegraphics[width=.48\textwidth,height=\textheight,keepaspectratio]{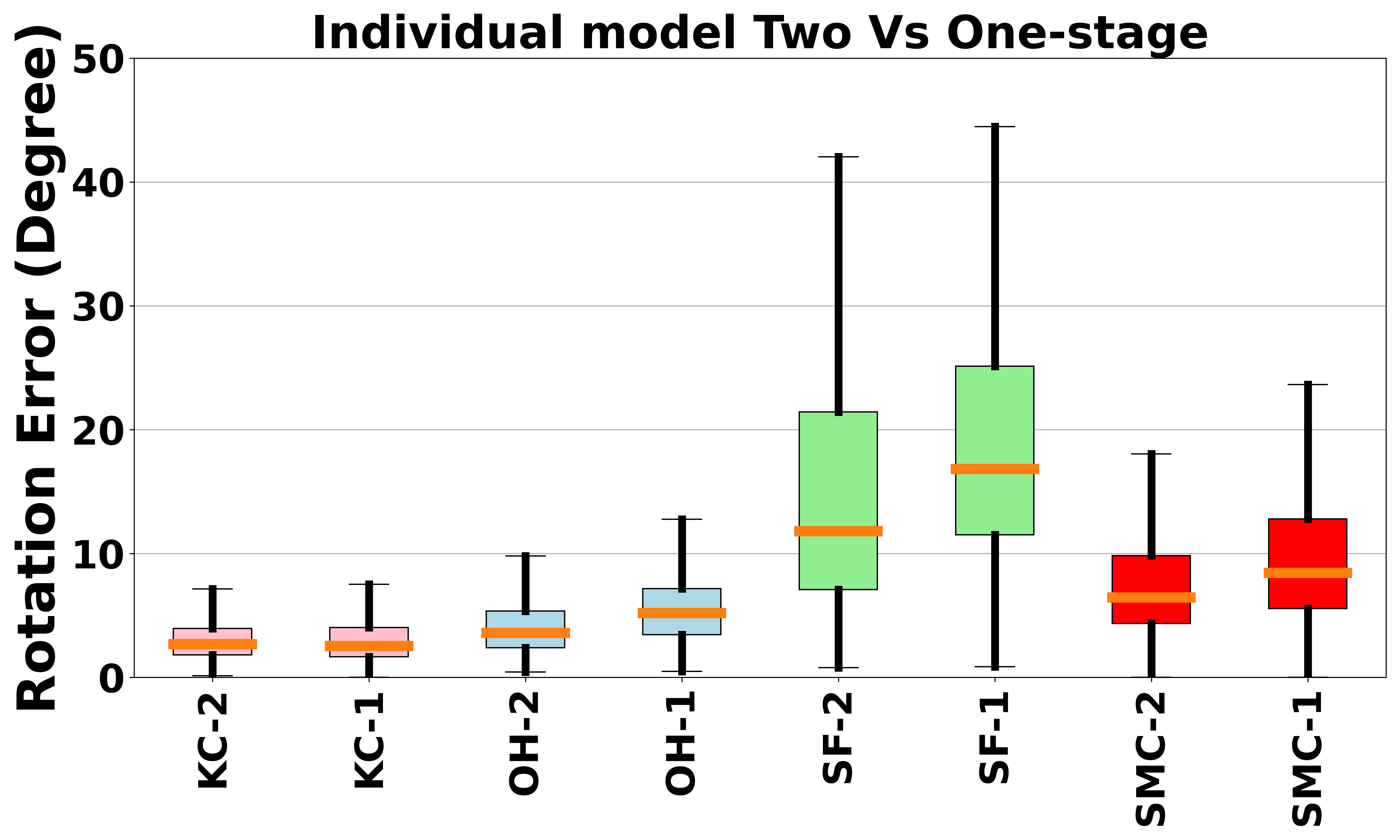}\label{ir}} \\
     \subfloat[]{\includegraphics[width=.48\textwidth,height=\textheight,keepaspectratio]{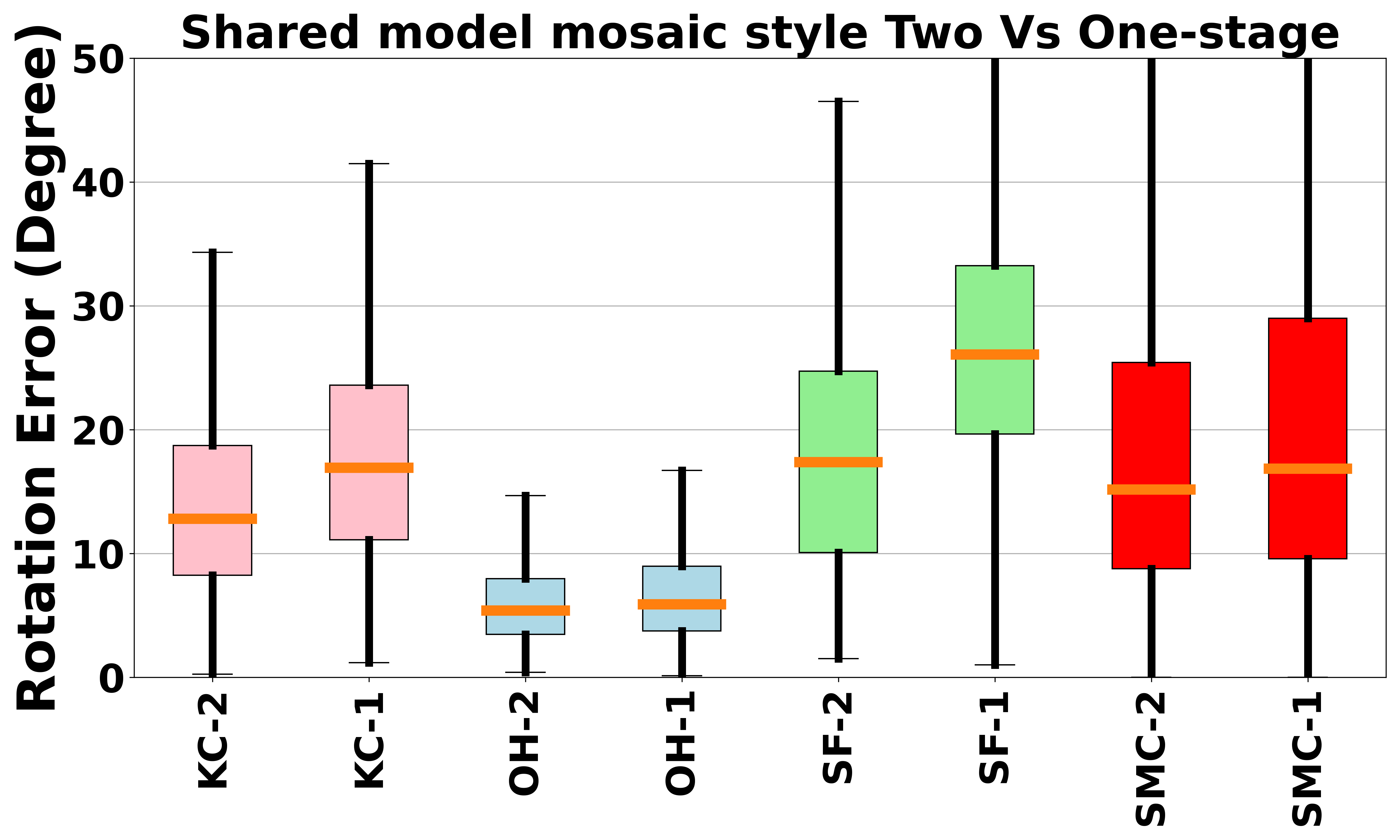}\label{ssr}}
     \subfloat[]{\includegraphics[width=.48\textwidth,height=\textheight,keepaspectratio]{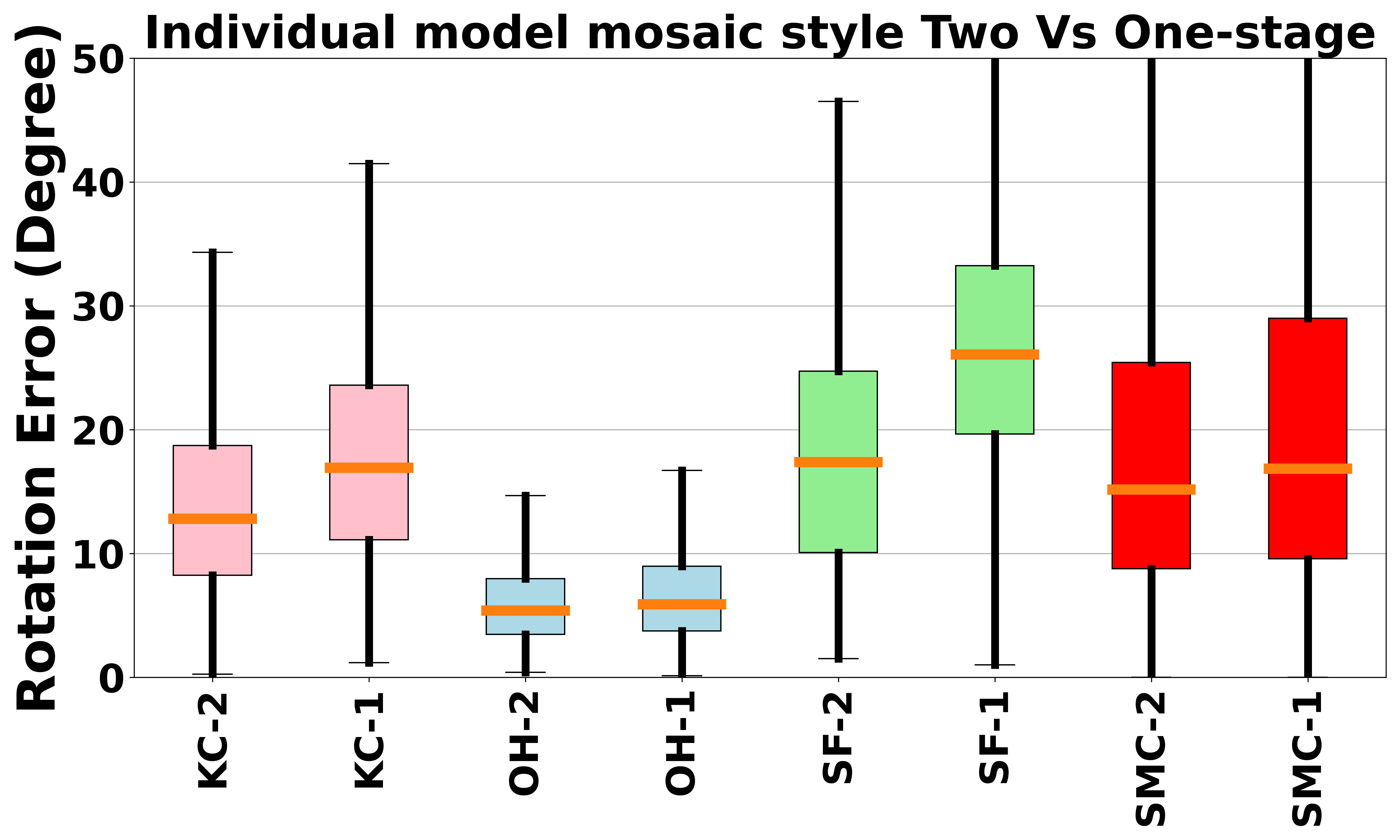}\label{sir}}
     \setlength{\belowcaptionskip}{-8pt}
     \caption{Box plot depicting rotation error distribution on test set. Less spread of box shows less variance and low central point of box (i.e. centroid) shows less bias. For better result low bias and variance is desired. (a) Shared model's rotation estimation with real data (b) Individual model's rotation estimation with real data (c) Shared model's rotation estimation with mosaic style data (d) Shared model's rotation estimation with mosaic style data. It should be noted that style transferred images are only used at test time. (
     \textbf{Notation}: KC - \textit{Kings College}, OH - \textit{Old Hospital}, SF - \textit{Shop Facade}, SMC - \textit{St Marys Church}, \textit{-2} $\rightarrow$ Two-Stage model results, \textit{-1} $\rightarrow$ One-Stage model results)
}
    \label{fig:boxplots}
\end{figure}

\begin{figure}[tb]
     \centering
     \subfloat[][Real data]{\includegraphics[width=.24\textwidth,height=\textheight,keepaspectratio]{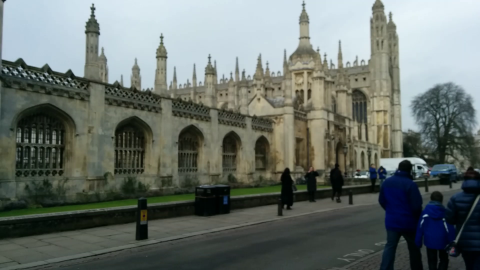}\label{rd}} \thinspace
     \subfloat[][Mosaic style]{\includegraphics[width=.24\textwidth,height=\textheight,keepaspectratio]{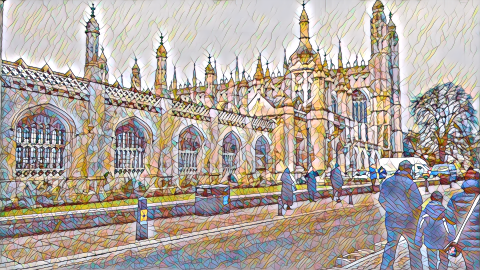}\label{md}} \thinspace
     \subfloat[][Starry style]{\includegraphics[width=.24\textwidth,height=\textheight,keepaspectratio]{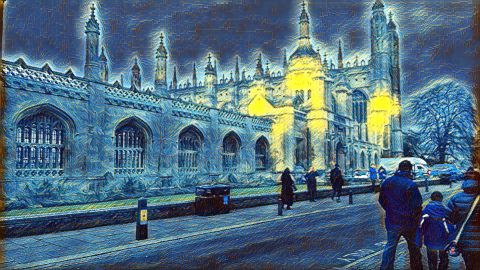}\label{sd}} \thinspace
     \subfloat[][Udnie style]{\includegraphics[width=.24\textwidth,height=\textheight,keepaspectratio]{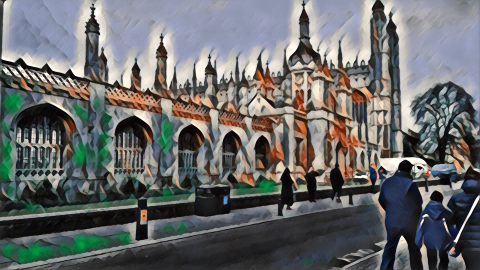}\label{ud}}
     \setlength{\belowcaptionskip}{-4pt}
     \caption{Sample of Real data along with styled transferred images with a generative adversarial network \cite{rusty2018faststyletransfer}}
     \label{fig:styleimages}
\end{figure}

\subsection{Two-stage vs One-stage approach}

For showing the effectiveness of two-stage training on improving rotation vector prediction, we ran an ablation study comparing one-stage training and two-stage training. Thus One-stage models are directly trained with the unit quaternion and an unnormalized translation vector as the ground truth (i.e, \textit{second set} ground truth labels) for 50 epochs. The results (in Table~\ref{table:perf2}) of models trained for each respective scene as well as all shared scene are shown in our ablations. When comparing our two-stage and one-stage models, two-stage training improves rotation prediction substantially while having little impact on translation prediction. Further, to bolster the claim that two-stage trained models perform better in different light (texture) conditions, we only augmented our test sets with three different style transferred images using \cite{rusty2018faststyletransfer} as shown in Fig.~\ref{fig:styleimages}. The inference results of translation and rotation vectors for one-stage training and two-stage training of each type of styled images are enlisted in Table~\ref{table:perf2}. Also, for real scene as well as for one style transferred scene the visual error box-plot are shown in Fig.~\ref{fig:boxplots}. Inference results indicate that two-stage training effectively strikes a balance, by also lowering the rotation error in augmented styled transferred images. Hence, two-stage training can be an effective alternative to handle textural variation as well as being independent of hyperparameter tuning for the translation and rotation losses during training.

\begin{table}[tb]
\begin{center}
\setlength{\belowcaptionskip}{-6pt}
\caption{Performance comparison of one-stage training and two-stage training models with inference results of translation and rotation vector. For “Real Data” both models(One-stage and Two-stage) are trained and then tested on respective real scene images.  However, for other three styled images, real data models are used for inference. \textbf{Bold} numbers represent occurrences where rotation prediction is better than one-stage models. Note that for Two-stage models (trained with \textit{first set}, and \textit{second set} ground truths) there are 4 Individual scene models, and 1 Shared scene model. Similarly, for One-stage models (directly trained with \textit{second set} ground truths) there are 4 Individual scene models, and 1 Shared scene model
}
\label{table:perf2}
\resizebox{\columnwidth}{!}{\begin{tabular}{c|c|c|c|c|c|c|c|c|c|}
  \hline
  \multicolumn{2}{c|}{\multirow{3}{*}{DATA \ APPROACH }} & \multicolumn{4}{c|}{Individual scene model} & \multicolumn{4}{c|}{Shared scene model} \\\cline{3-10}
  \multicolumn{2}{c|}{} & \multicolumn{2}{c|}{Two-stage} & \multicolumn{2}{c|}{One-stage} & \multicolumn{2}{c|}{Two-stage} & \multicolumn{2}{c|}{One-stage} \\\cline{3-10}

  \multicolumn{2}{c|}{} & Rotation $^{\circ}$ & Translation m & Rotation $^{\circ}$ & Translation m & Rotation $^{\circ}$ & Translation m & Rotation $^{\circ}$ & Translation m \\
  \hline
  \multirow{4}{*}{ Real data } & King’s C. & 2.70 & 1.45 & 2.54 & 1.35 & \textbf{2.93} & 1.51 & 4.21 & 1.30 \\
  & Old Hospital & \textbf{3.60} & 2.51 & 5.20 & 2.70 & \textbf{3.63} & 2.24 & 4.45 & 2.07 \\
  & Shop Facade & \textbf{11.80} & 2.63 & 16.85 & 2.80 & \textbf{5.63} & 1.34 & 5.88 & 1.28 \\
  & St Mary’s C. & \textbf{6.47} & 2.14 & 8.44 & 2.19 & \textbf{6.30} & 2.31 & 6.84 & 2.01 \\\cline{1-10}
  \multirow{4}{*}{ Mosaic styled data } & King’s C. & \textbf{12.82} & 8.82 & 16.94 & 8.06 & \textbf{10.16} & 9.07 & 10.81 & 8.42 \\
  & Old Hospital & \textbf{5.42} & 4.40 & 5.92 & 4.14 & 8.93 & 5.96 & 8.24 & 5.24 \\
  & Shop Facade & \textbf{17.40} & 4.71 & 26.08 & 4.86 & \textbf{14.63} & 3.57 & 12.56 & 4.37 \\
  & St Mary’s C. & \textbf{15.19} & 5.08 & 16.87 & 6.16 & 17.84 & 7.07 & 16.75 & 6.97 \\\cline{1-10}
  \multirow{4}{*}{ Starry styled data } & King’s C. & \textbf{17.38} & 9.20 & 27.34 & 8.96 & 12.80 & 9.62 & 11.26 & 9.64 \\
  & Old Hospital & \textbf{6.37} & 4.70 & 6.49 & 4.24 & 10.33 & 6.33 & 8.47 & 5.66 \\
  & Shop Facade & \textbf{21.96} & 5.76 & 28.10 & 5.24 & 18.14 & 4.74 & 15.83 & 5.40 \\
  & St Mary’s C. & 18.25 & 6.00 & 17.35 & 6.04 & 20.21 & 7.44 & 18.91 & 7.00 \\\cline{1-10}
  \multirow{4}{*}{ Udnie styled data } & King’s C. & \textbf{6.63} & 5.15 & 6.78 & 4.96 & \textbf{8.78} & 6.98 & 9.35 & 6.24 \\
  & Old Hospital & \textbf{4.97} & 3.81 & 5.48 & 3.68 & \textbf{5.89} & 3.92 & 6.63 & 3.62 \\
  & Shop Facade & \textbf{15.49} & 3.73 & 17.46 & 3.20 & \textbf{9.27} & 2.93 & 10.46 & 2.89 \\
  & St Mary’s C. & \textbf{13.99} & 4.61 & 15.18 & 4.59 & \textbf{14.22} & 5.45 & 14.72 & 4.97 \\\cline{1-10}
\end{tabular}}
\end{center}
\end{table}

\subsection{Secondary Data Collection and Evaluation}
\begin{figure}[tb]
     \centering
     \subfloat[][]{\includegraphics[width=.23\textwidth,height=\textheight,keepaspectratio]{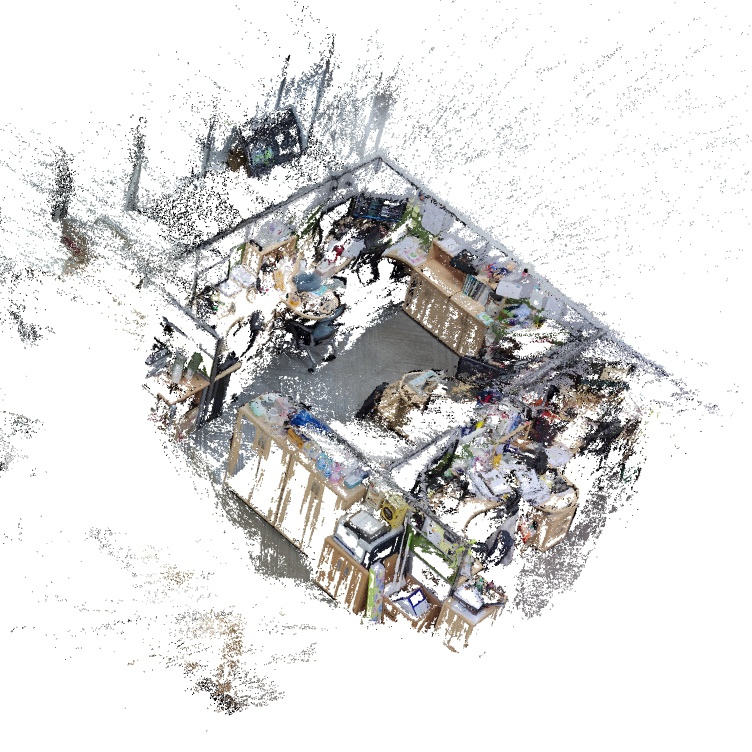}\label{td}} \thinspace \thinspace
     \subfloat[][]{\includegraphics[width=.23\textwidth,height=\textheight,keepaspectratio]{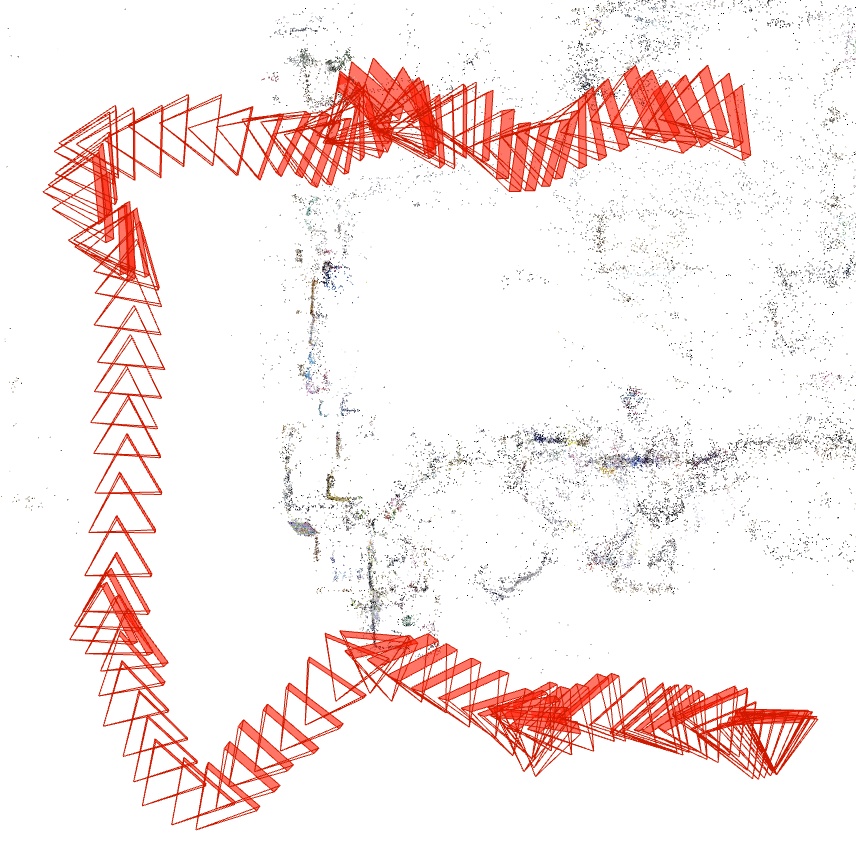}\label{ts}} \thinspace \thinspace
     \subfloat[][]{\includegraphics[width=.23\textwidth,height=\textheight,keepaspectratio]{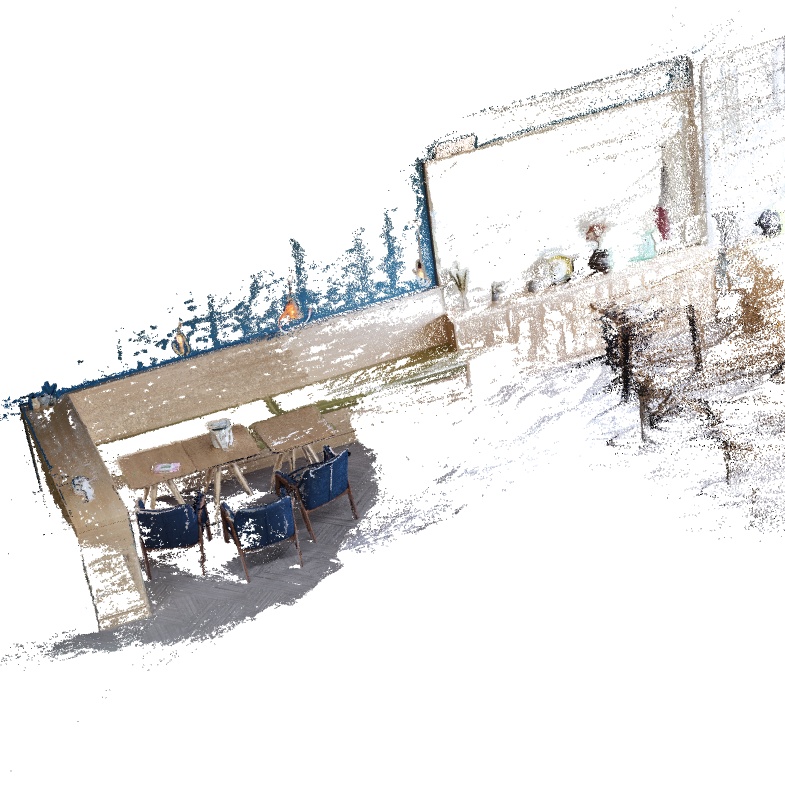}\label{ted}} \thinspace \thinspace
     \subfloat[][]{\includegraphics[width=.23\textwidth,height=\textheight,keepaspectratio]{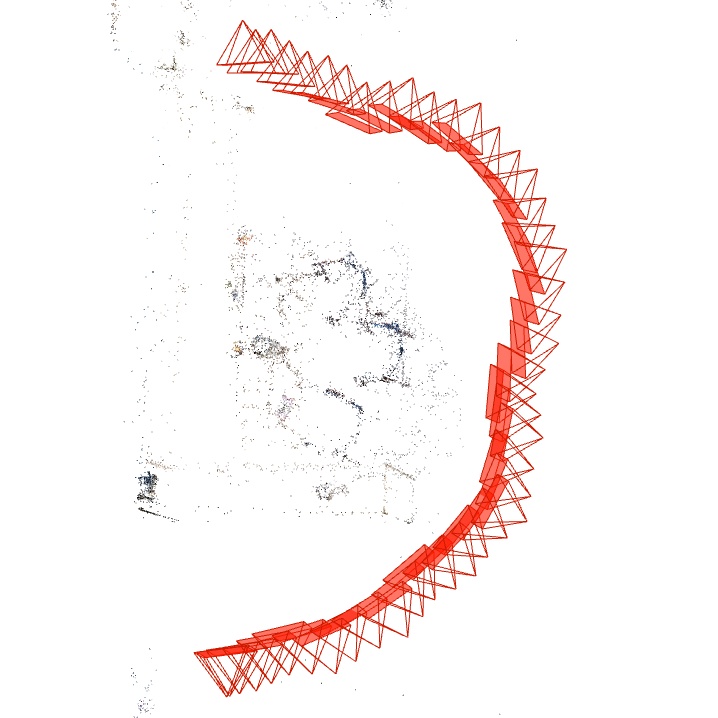}\label{tes}}
     \setlength{\belowcaptionskip}{-8pt}
     \caption{(a,b) Training sequence dense and sparse reconstruction (c,d) Testing sequence dense and sparse reconstruction, for data collected with mobile camera visualized with COLMAP \cite{schoenberger2016sfm}}
     \label{fig:newdata}
\end{figure}

To further evaluate the relative pose estimation of our method we collected auxiliary data. Video is recorded inside of the lab space and common workspace for training and testing respectively. We capture the video using a mobile(Samsung S22) camera at the resolution of 1980x1080 at 30 frames per second. Video is then sampled at 2 hertz to obtain the images at sequential time steps. With the sampled images we use COLMAP \cite{schoenberger2016sfm} to obtain the sparse, dense reconstruction and pose information. We display the training and test sequence we collected with the mobile camera in Fig.~\ref{fig:newdata}.

The reconstructed pose information from COLMAP \cite{schoenberger2016sfm} is specified as the projection from world to the camera coordinate system of an image with quaternion \((q'w, \thinspace q'x, \thinspace q'y, \thinspace q'z)\) and translation vector \( t' = (x', \thinspace y', \thinspace z') \). We sort the poses in the ascending order of time stamp, and acquire the corresponding rotation matrices $ R'\in \mathbb{R}^{3X3} $ using pyquaternion \cite{pyquaternion}. Actual camera trajectory's translation is calculated as.

\begin{equation}
t = R'^{ \thinspace -1 } \thinspace t'
\end{equation}  

However, COLMAP provides the pose where the origin of the pose is not necessarily in reference to the first frame of image, i.e. as part of bundle adjustment process it drifts away from the first frame coordinate system\footnote{https://github.com/colmap/colmap/issues/1229}. In order to correct for this we further calculate the transformations for the rotation matrices and translation vectors by using the first frame's pose \(R_{ff}',\thinspace t_{ff}\) as follows.
\begin{equation}
R_{TRF} = R_{ff}'^{ \thinspace -1 }
\end{equation}  
\begin{equation}
t_{TRF} = -t_{ff}
\end{equation}  

\(R_{TRF},\thinspace t_{TRF}\) is then used transform all the pose in the dataset like the following,
\begin{equation}
R_{(abs)_i} = R_{i}'\thinspace R_{TRF}\qquad \forall i \in N
\end{equation}
\begin{equation}
t_{(abs)_i} = t_{i}\thinspace + t_{TRF}\qquad \forall i \in N 
\end{equation}
where N is a total number of images in the dataset. The above formulation enforces a coordinate system referenced with the first frame as the origin. i.e. for the first frame, rotation \(R_{(abs)_1} = 3X3\) as an identity matrix and translation vector \(t_{(abs)_1} = [0, \thinspace 0, \thinspace 0]^T\). After obtaining the appropriate absolute poses for all time step images, rigid transformation $ T_{abs}\in \mathbb{R}^{4X4} $ encapsulating rotation and translation can be calculated as.
\begin{equation}
T_{(abs)_i} =  \begin{bmatrix}
R_{{abs}_i} & t_{(abs)_i}\\
0 & 1
\end{bmatrix} \qquad \forall i \in N 
\end{equation}
Finally, the relative pose for the current frame in reference to the previous frame's pose is calculated as.
\begin{equation}
T_{(rel)_i} =  \begin{bmatrix}
R_{{abs}_i} & t_{(abs)_i}\\
0 & 1
\end{bmatrix} \begin{bmatrix}
R_{{abs}_{i-1}} & t_{(abs)_{i-1}}\\
0 & 1
\end{bmatrix}^{-1} \quad \forall i \in N \quad where \quad i\neq 1
\end{equation}
The rotation matrix can be converted back to a quaternion representation. We sampled 117 image frames as our training set corresponding to a total of 116 image pairs and sampled 46 image frames as our test set corresponding to a total 45 image pairs. This training set was then used for fine tuning the shared two-stage model for additional 30 epochs. The resulting model was tested on the unseen test set, in which a median error of 6.00$^{\circ}$ and 0.64m were obtained for rotation and translation, respectively.

\section{Conclusion}

This paper investigates the problem of finding camera relative pose for various scene domains. The proposed siamese architecture using a MobileNetV3-Large backbone is lightweight as well as achieving comparative performance when compared to other computationally heavy architectures. To alleviate the need of a hyperparameter to balance between translation and rotation losses, a two-stage training is proposed. Experiments show that a two-stage models trained on multiple scene domains improve the generalization of the network when tested against style transferred augmented images. The two-stage training process not only improve the convergence speed of the model but also remove the need for a hyperparameter to weigh between translation and rotation losses. When compared with baseline methods, the proposed method improves translation estimation while achieving comparable rotation estimation.

\bibliographystyle{splncs04}
\bibliography{reference}

\begin{thebibliography}{10}
\providecommand{\url}[1]{\texttt{#1}}
\providecommand{\urlprefix}{URL }
\providecommand{\doi}[1]{https://doi.org/#1}

\bibitem{bailo2018efficient}
Bailo, O., Rameau, F., Joo, K., Park, J., Bogdan, O., Kweon, I.S.: Efficient
  adaptive non-maximal suppression algorithms for homogeneous spatial keypoint
  distribution. Pattern Recognition Letters  \textbf{106},  53--60 (2018)

\bibitem{bay2008speeded}
Bay, H., Ess, A., Tuytelaars, T., Van~Gool, L.: Speeded-up robust features
  (surf). Computer vision and image understanding  \textbf{110}(3),  346--359
  (2008)

\bibitem{brachmann2017dsac}
Brachmann, E., Krull, A., Nowozin, S., Shotton, J., Michel, F., Gumhold, S.,
  Rother, C.: Dsac-differentiable ransac for camera localization. In:
  Proceedings of the IEEE conference on computer vision and pattern
  recognition. pp. 6684--6692 (2017)

\bibitem{opencv_library}
Bradski, G.: {The OpenCV Library}. Dr. Dobb's Journal of Software Tools  (2000)

\bibitem{chen2021wide}
Chen, K., Snavely, N., Makadia, A.: Wide-baseline relative camera pose
  estimation with directional learning. In: Proceedings of the IEEE/CVF
  Conference on Computer Vision and Pattern Recognition. pp. 3258--3268 (2021)

\bibitem{dusmanu2019d2}
Dusmanu, M., Rocco, I., Pajdla, T., Pollefeys, M., Sivic, J., Torii, A.,
  Sattler, T.: D2-net: A trainable cnn for joint detection and description of
  local features. arXiv preprint arXiv:1905.03561  (2019)

\bibitem{en2018rpnet}
En, S., Lechervy, A., Jurie, F.: Rpnet: An end-to-end network for relative
  camera pose estimation. In: Proceedings of the European Conference on
  Computer Vision (ECCV) Workshops. pp.~0--0 (2018)

\bibitem{fischler1981random}
Fischler, M.A., Bolles, R.C.: Random sample consensus: a paradigm for model
  fitting with applications to image analysis and automated cartography.
  Communications of the ACM  \textbf{24}(6),  381--395 (1981)

\bibitem{graziani2021scale}
Graziani, M., Lompech, T., M{\"u}ller, H., Depeursinge, A., Andrearczyk, V.: On
  the scale invariance in state of the art cnns trained on imagenet. Machine
  Learning and Knowledge Extraction  \textbf{3}(2),  374--391 (2021)

\bibitem{hartley2013multiple}
Hartley, R., Zisserman, A.: Multiple view geometry in computer vision
  (cambridge university, 2003). C1 C3  \textbf{2} (2013)

\bibitem{hartley1997defense}
Hartley, R.I.: In defense of the eight-point algorithm. IEEE Transactions on
  pattern analysis and machine intelligence  \textbf{19}(6),  580--593 (1997)

\bibitem{howard2019searching}
Howard, A., Sandler, M., Chu, G., Chen, L.C., Chen, B., Tan, M., Wang, W., Zhu,
  Y., Pang, R., Vasudevan, V., et~al.: Searching for mobilenetv3. In:
  Proceedings of the IEEE/CVF international conference on computer vision. pp.
  1314--1324 (2019)

\bibitem{hwang2018ferrite}
Hwang, K., Cho, J., Park, J., Har, D., Ahn, S.: Ferrite position identification
  system operating with wireless power transfer for intelligent train position
  detection. IEEE Transactions on Intelligent Transportation Systems
  \textbf{20}(1),  374--382 (2018)

\bibitem{kendall2017geometric}
Kendall, A., Cipolla, R.: Geometric loss functions for camera pose regression
  with deep learning. In: Proceedings of the IEEE conference on computer vision
  and pattern recognition. pp. 5974--5983 (2017)

\bibitem{kendall2015posenet}
Kendall, A., Grimes, M., Cipolla, R.: Posenet: A convolutional network for
  real-time 6-dof camera relocalization. In: Proceedings of the IEEE
  international conference on computer vision. pp. 2938--2946 (2015)

\bibitem{kim2020pose}
Kim, S., Kim, I., Vecchietti, L.F., Har, D.: Pose estimation utilizing a gated
  recurrent unit network for visual localization. Applied Sciences
  \textbf{10}(24), ~8876 (2020)

\bibitem{kingma2014adam}
Kingma, D.P., Ba, J.: Adam: A method for stochastic optimization. arXiv
  preprint arXiv:1412.6980  (2014)

\bibitem{lee2019optimal}
Lee, S., Lee, J., Jung, H., Cho, J., Hong, J., Lee, S., Har, D.: Optimal power
  management for nanogrids based on technical information of electric
  appliances. Energy and Buildings  \textbf{191},  174--186 (2019)

\bibitem{lowe2004distinctive}
Lowe, D.G.: Distinctive image features from scale-invariant keypoints.
  International journal of computer vision  \textbf{60}(2),  91--110 (2004)

\bibitem{melekhov2017relative}
Melekhov, I., Ylioinas, J., Kannala, J., Rahtu, E.: Relative camera pose
  estimation using convolutional neural networks. In: International Conference
  on Advanced Concepts for Intelligent Vision Systems. pp. 675--687. Springer
  (2017)

\bibitem{rusty2018faststyletransfer}
Mina, R.: fast-neural-style: Fast style transfer in pytorch!
  \url{https://github.com/iamRusty/fast-neural-style-pytorch} (2018)

\bibitem{moraes2017distributed}
Moraes, C., Myung, S., Lee, S., Har, D.: Distributed sensor nodes charged by
  mobile charger with directional antenna and by energy trading for balancing.
  Sensors  \textbf{17}(1), ~122 (2017)

\bibitem{nister2004efficient}
Nist{\'e}r, D.: An efficient solution to the five-point relative pose problem.
  IEEE transactions on pattern analysis and machine intelligence
  \textbf{26}(6),  756--770 (2004)

\bibitem{adaptiveavgpool2d-pytorch1.12documentation}
Paszke, A., Gross, S., Massa, F., Lerer, A., Bradbury, J., Chanan, G., Killeen,
  T., Lin, Z., Gimelshein, N., Antiga, L., Desmaison, A., Kopf, A., Yang, E.,
  DeVito, Z., Raison, M., Tejani, A., Chilamkurthy, S., Steiner, B., Fang, L.,
  Bai, J., Chintala, S.: Adaptiveavgpool2d¶,
  \url{https://pytorch.org/docs/stable/generated/torch.nn.AdaptiveAvgPool2d.html}

\bibitem{NEURIPS2019_9015}
Paszke, A., Gross, S., Massa, F., Lerer, A., Bradbury, J., Chanan, G., Killeen,
  T., Lin, Z., Gimelshein, N., Antiga, L., Desmaison, A., Kopf, A., Yang, E.,
  DeVito, Z., Raison, M., Tejani, A., Chilamkurthy, S., Steiner, B., Fang, L.,
  Bai, J., Chintala, S.: Pytorch: An imperative style, high-performance deep
  learning library. In: Wallach, H., Larochelle, H., Beygelzimer, A.,
  d\textquotesingle Alch\'{e}-Buc, F., Fox, E., Garnett, R. (eds.) Advances in
  Neural Information Processing Systems 32, pp. 8024--8035. Curran Associates,
  Inc. (2019),
  \url{http://papers.neurips.cc/paper/9015-pytorch-an-imperative-style-high-performance-deep-learning-library.pdf}

\bibitem{philbin2007object}
Philbin, J., Chum, O., Isard, M., Sivic, J., Zisserman, A.: Object retrieval
  with large vocabularies and fast spatial matching. In: 2007 IEEE conference
  on computer vision and pattern recognition. pp.~1--8. IEEE (2007)

\bibitem{poursaeed2018deep}
Poursaeed, O., Yang, G., Prakash, A., Fang, Q., Jiang, H., Hariharan, B.,
  Belongie, S.: Deep fundamental matrix estimation without correspondences. In:
  Proceedings of the European Conference on Computer Vision (ECCV) Workshops.
  pp.~0--0 (2018)

\bibitem{raguram2008comparative}
Raguram, R., Frahm, J.M., Pollefeys, M.: A comparative analysis of ransac
  techniques leading to adaptive real-time random sample consensus. In:
  European conference on computer vision. pp. 500--513. Springer (2008)

\bibitem{rublee2011orb}
Rublee, E., Rabaud, V., Konolige, K., Bradski, G.: Orb: An efficient
  alternative to sift or surf. In: 2011 International conference on computer
  vision. pp. 2564--2571. Ieee (2011)

\bibitem{sarlin2020superglue}
Sarlin, P.E., DeTone, D., Malisiewicz, T., Rabinovich, A.: Superglue: Learning
  feature matching with graph neural networks. In: Proceedings of the IEEE/CVF
  conference on computer vision and pattern recognition. pp. 4938--4947 (2020)

\bibitem{schoenberger2016sfm}
Sch\"{o}nberger, J.L., Frahm, J.M.: Structure-from-motion revisited. In:
  Conference on Computer Vision and Pattern Recognition (CVPR) (2016)

\bibitem{seo2019rewards}
Seo, M., Vecchietti, L.F., Lee, S., Har, D.: Rewards prediction-based credit
  assignment for reinforcement learning with sparse binary rewards. IEEE Access
   \textbf{7},  118776--118791 (2019)

\bibitem{sun2021loftr}
Sun, J., Shen, Z., Wang, Y., Bao, H., Zhou, X.: Loftr: Detector-free local
  feature matching with transformers. In: Proceedings of the IEEE/CVF
  conference on computer vision and pattern recognition. pp. 8922--8931 (2021)

\bibitem{pyquaternion}
Wynn, K.: pyquaternion. \url{https://github.com/KieranWynn/pyquaternion} (2020)

\bibitem{yang2020rcpnet}
Yang, C., Liu, Y., Zell, A.: Rcpnet: Deep-learning based relative camera pose
  estimation for uavs. In: 2020 International Conference on Unmanned Aircraft
  Systems (ICUAS). pp. 1085--1092. IEEE (2020)

\bibitem{yew2022regtr}
Yew, Z.J., Lee, G.H.: Regtr: End-to-end point cloud correspondences with
  transformers. In: Proceedings of the IEEE/CVF Conference on Computer Vision
  and Pattern Recognition. pp. 6677--6686 (2022)

\end{thebibliography}
\end{document}